\documentclass[10pt]{article} 
\usepackage[preprint]{rlc}
\usepackage[T1]{fontenc}

\usepackage{amssymb}            
\usepackage{mathtools}          
\usepackage{mathrsfs}           
\usepackage{graphicx}           
\usepackage{subcaption}         
\usepackage[space]{grffile}     
\usepackage{url}                

\usepackage{multirow}
\usepackage{amsmath}
\usepackage{bm}
\usepackage{algorithm}
\usepackage{algpseudocode}
\usepackage{subcaption}
\usepackage{booktabs}

\title{Semi-Supervised One-Shot Imitation Learning}

\author{
Philipp Wu$^{*}$ \\ 
University of California, Berkeley
\And
Kourosh Hakhamaneshi$^{*}$ \\
University of California, Berkeley
\And
Yuqing Du \\
University of California, Berkeley
\And
Igor Mordatch \\
Google Deepmind
\And
Aravind Rajeswaran \\
University of California, Berkeley
\And
Pieter Abbeel \\
University of California, Berkeley
}

\begin{document}

\maketitle

\begin{abstract}
One-shot Imitation Learning~(OSIL) aims to imbue AI agents with the ability to learn a new task from a single demonstration. To supervise the learning, OSIL typically requires a prohibitively large number of paired expert demonstrations -- i.e. trajectories corresponding to different variations of the same semantic task. To overcome this limitation, we introduce the semi-supervised OSIL problem setting, where the learning agent is presented with a large dataset of trajectories with no task labels (i.e. an unpaired dataset), along with a small dataset of multiple demonstrations per semantic task (i.e. a paired dataset). 
This presents a more realistic and practical embodiment of few-shot learning and requires the agent to effectively leverage weak supervision from a large dataset of trajectories.
Subsequently, we develop an algorithm specifically applicable to this semi-supervised OSIL setting. Our approach first learns an embedding space where different tasks cluster uniquely. We utilize this embedding space and the clustering it supports to self-generate pairings between trajectories in the large unpaired dataset. Through empirical results on simulated control tasks, we demonstrate that OSIL models trained on such self-generated pairings are competitive with OSIL models trained with ground-truth labels, presenting a major advancement in the label-efficiency of OSIL.
\end{abstract}

\section{Introduction}
\label{sec:intro}

Humans are capable of learning new tasks and behaviors by imitating others we observe. Furthermore, we are remarkably data efficient, often requiring just a single demonstration. One-shot imitation learning (OSIL)~\citep{rocky_osil} aims to imbue AI agents with similar capabilities. It takes a meta-learning~\citep{schmidhuber1987, naik, thrun} approach and considers several paired demonstrations -- i.e. expert trajectories corresponding to different variations of the semantic task. OSIL learns to reconstruct one trajectory by conditioning on its paired trajectory, implicitly capturing the task semantics. At test time, the resulting agent can directly complete a new task by conditioning on a demonstration of the said task.
However, this method often requires prohibitively large amounts of paired trajectories such that the agent experiences enough task variations in diverse environment instantiations to learn a generalizable policy.
Collecting such a dataset of demonstrations can be prohibitively expensive, requiring significant engineering effort and/or human data annotation time. In order to improve the data efficiency of OSIL, and expand its applicability, we introduce and study a semi-supervised paradigm for OSIL.

In recent years, we have seen an increase in our ability to collect unsupervised trajectory data in several applications including robotics. This includes access to historical offline datasets~\citep{levine2020offline, fu2020d4rl, Gulcehre2020RLUB}, teleoperation and play data in virtual reality~\citep{Rajeswaran_RSS_18, lynch2019play, gupta2019rpl}, and reward-free exploration~\citep{pathak2017curiosity, eysenbach2018diversity, liu2021aps}. Our goal is to leverage these large, abundant, but unlabelled datasets to create a more scalable pathway for OSIL. 
A direct and naive application of OSIL would require humans to manually annotate these datasets with semantic task descriptions, or manually pair together similar trajectories, which can be expensive and time consuming. We draw inspiration from semi-supervised learning~\citep{vanEngelen2019ASO} in computer vision and natural language processing (NLP), which has emerged as a dominant paradigm to utilize a small labeled dataset in conjunction with large quantities of unlabeled data to train high-quality models \citep{pseudo_labeling, xie2020self, chen2020simclr, devlin2018bert}. Analogously, we aim to bring the power of semi-supervised learning to OSIL by learning from both task-agnostic and unlabelled trajectories as well as a small dataset of annotated (paired) trajectories.

\begin{figure}
\centering
\begin{subfigure}[t]{0.39\textwidth}
    \includegraphics[width=\textwidth]{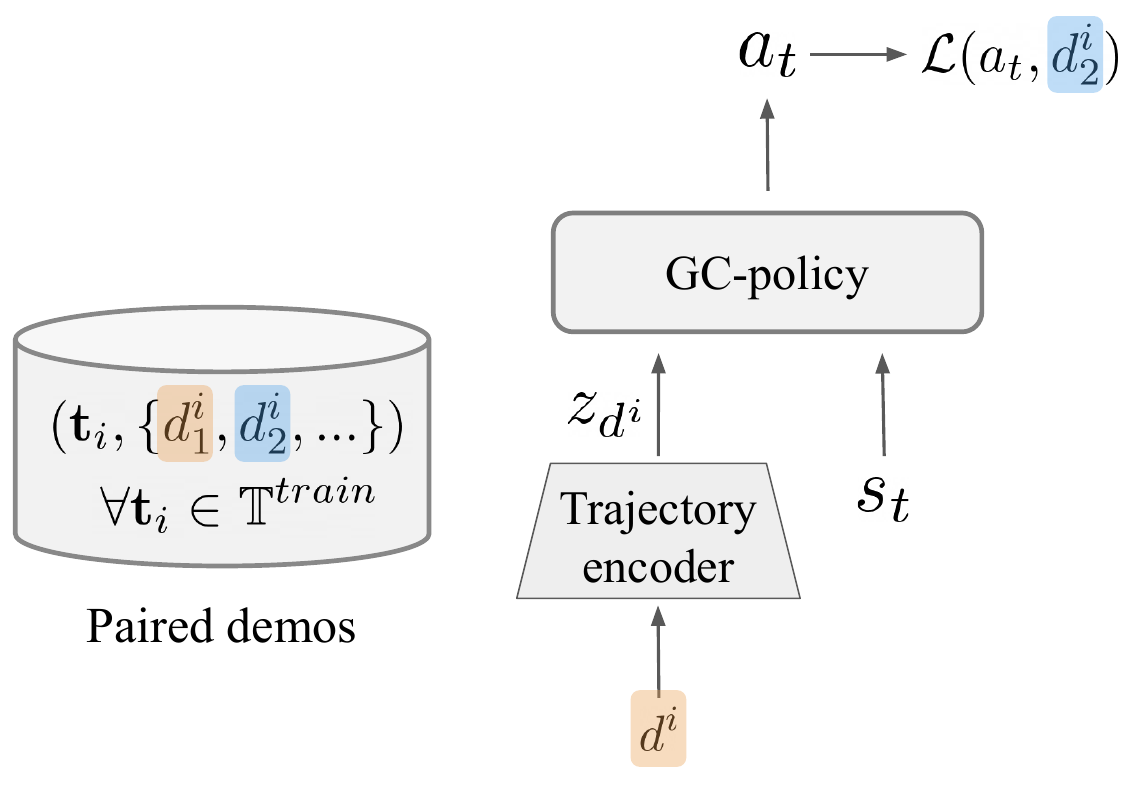}
    \caption{Supervised OSIL setting.}
    \label{suposilsub}
\end{subfigure}
\hfill
\begin{subfigure}[t]{0.6\textwidth}
    \includegraphics[width=\textwidth]{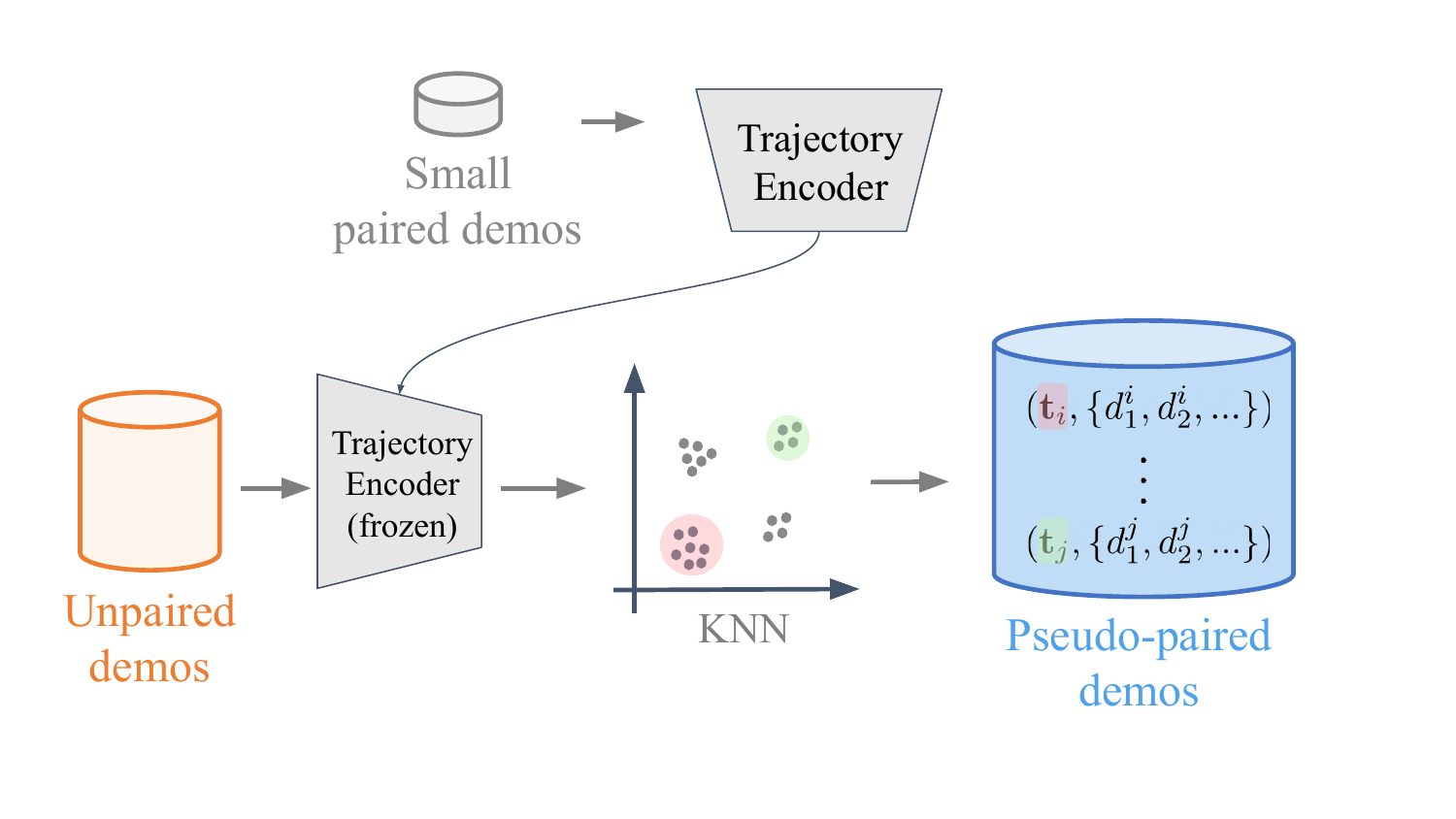}
    \caption{Our semi-supervised OSIL approach.}
    \label{ssosilsub}
\end{subfigure}
\caption{(Left) Depiction of the supervised (classical) OSIL setting, where the encoder and policy are trained using several trajectories ($d$) sharing the same task label ($\bm{t}$). (Right) Our semi-supervised OSIL setting instead requires only a large unlabelled dataset of trajectories, and a small paired dataset. For our method, a teacher trajectory encoder is first trained using the labeled dataset. This encoder is then used to construct a pseudo-paired trajectory set by retrieving the $k$ nearest neighbors of each trajectory. We can then train a student on this pseudo-labeled dataset, as in supervised OSIL. Optionally, this relabelling and training procedure can be repeated iteratively.}
\label{fig:ssosil_overview}
\vspace*{-10pt}
\end{figure}

Our algorithmic approach to semi-supervised OSIL is based on self-training~\citep{Triguero2013SelflabeledTF, Yarowsky1995UnsupervisedWS, xie2020self}, a prominent approach to semi-supervised learning. In self-training, a teacher network is first trained on a small labeled dataset, and then used to provide pseudo-labels for a larger unlabeled dataset \citep{teacher_student}. This process is repeated multiple times to progressively learn higher quality labels for the entire dataset, ultimately training models with competitive performance despite considerably reduced data annotation effort. 
To adapt this self-training approach to semi-supervised OSIL, we start with training a teacher encoder-decoder architecture in the standard supervised OSIL fashion, as illustrated in Figure \ref{fig:ssosil_overview} (a), with the available paired dataset.
We show that even when the teacher does not reach a high task success, the embedding space is sufficiently structured to distinctly cluster different semantic tasks, enabling the generation of pseduo-labelled pairings between nearest neighbors in the embedding space.
By bootstrapping on the pseudo-labels obtained from the trajectory clusters in embedding space, we can train a student architecture that outperforms the teacher.

\textbf{Our Contributions} in this work are summarized below.
\vspace*{-5pt}
\begin{enumerate}
    \itemsep0em
    \item We introduce and formalize the semi-supervised OSIL setting.
    \item We propose a novel label-efficient student-teacher trajectory relabeling approach for semi-supervised OSIL that extends the ideas of self-training and distillation from CV and NLP.
    \item In a semantic goal navigation task, we find that our method enables an agent trained with only 15\% of labelled data to match a fully supervised agent. In a sequential goal navigation task our method approaches fully supervised performance with only 5\% of labelled data. 
    \item We ablate each component of our method, demonstrating their importance to the overall algorithmic contribution.
\end{enumerate}
\vspace*{-5pt}
After the anonymous review phase, we are committed to providing open source for the environments and experiments to facilitate reproducibility and future extensions.
\section{Related work}
\label{related_work}

\paragraph{One-Shot Imitation Learning (OSIL)}
The OSIL framework was originally introduced by \citet{rocky_osil} to endow AI agents with the capability to learn from a single demonstration. OSIL relies on access to ``paired'' demonstrations -- i.e. expert trajectories that correspond to different variations of the same semantic task. OSIL then learns by conditioning on one trajectory to reconstruct the paired demonstration, enabling it to implicitly learn the notion of task semantics. Through this view, OSIL has parallels to meta-learning or learning-to-learn~\citep{Ren2018MetaLearningFS, Finn2017ModelAgnosticMF, Vinyals2016MatchingNF, Chebotar2021MetaLV, Rajeswaran2019IMAML} as studied broadly in (supervised) machine learning and inverse RL~\citep{Das2020ModelBasedIR, Yu2019MetaInverseRL}.

Since the original work of \citet{rocky_osil}, OSIL has seen several extensions including extensions to visual observation spaces~\citep{finn_meta_osil}, improving task-level generalization~\citep{mandi_osil}, and architectural innovations like transformers~\citep{traj_osil}. Nevertheless, the need for a large number of paired demonstrations has limited the broad applicability of OSIL. Our work aims to improve this label efficiency of OSIL by also effectively utilizing a large number of unlabelled (i.e. unpaired) demonstrations, which are often substantially easier to obtain, for example through play data collection~\citep{lynch2019play}.

\paragraph{Semi-Supervised Learning}
The field of semi-supervised learning~\citep{Zhu2005SemiSupervisedLL} studies methods to simultaneously learn from large unlabelled datasets and small labelled datasets. 
Computer vision, NLP, and speech recognition have been exploring ways to utilize large unlabelled datasets scraped from the internet without expensive and time-intensive human annotations.
This has resulted in a wide array of approaches to semi-supervised learning~\citep{Zhu2005SemiSupervisedLL, vanEngelen2019ASO}. One dominant paradigm involves pre-training visual representations using unlabelled datasets followed by downstream supervised learning. The representations can be pre-trained with contrastive learning~\citep{Hjelm2019LearningDR, chen2020simclr}, generative modeling~\citep{Goodfellow2014GenerativeAN}, autoencoders~\citep{Vincent2008denoisingAE, He2021MaskedAA, wu2023mtm} and more. However, such representations lack knowledge of downstream task, and thus might be harder to train, require human priors like appropriate choice of augmentations, or demand very large quantities of unlabelled data.

An alternative and popular approach to semi-supervised learning is self-training~\citep{Triguero2013SelflabeledTF, Yarowsky1995UnsupervisedWS, xie2020self}, where a supervised ``teacher'' model is first trained on a small labelled dataset and used to generate pseudo-labels for the unsupervised dataset. Subsequently, a student model is trained on both the supervised dataset and the pseudo-labelled dataset. We refer readers to survey works~\citep{vanEngelen2019ASO} on semi-supervised learning for more discussion.
Our algorithmic approach to semi-supervised OSIL is closer to self-training, and thus has the advantage of being more task-directed in nature. We also perform contrastive representation learning as an auxiliary task and find that it plays an important role, but is insufficient by itself.

\paragraph{Semi-Supervised Learning in RL and IL} Improving label efficiency for policy learning, through approaches similar to semi-supervised learning, has been studied in other contexts like reward and goal labels. Prior works tackle the challenge of learning from data without reward/goal labels by either explicitly learning a reward function through inverse RL~\citep{Abbeel2004ApprenticeshipLV, Ziebart2008MaximumEI, Finn2016GuidedCL}, adversarial imitation learning~\citep{Ho2016GAIL, Fu2018AIRL, Rafailov2021VMAIL}, learning a reward/goal classifier~\citep{Fu2018VICE, Eysenbach2021Recursive}, or by simply assuming a pseudo baseline reward~\citep{Yu2022UDS}. In contrast to such prior work, we focus on improving the label efficiency of OSIL, where the need for a large number of paired demonstrations has limited real-world applicability. To our knowledge, our work is the first to study semi-supervised learning approaches to improve label efficiency for OSIL.
\section{Problem Formulation}
\label{problem_formulation}

Following \citet{rocky_osil}, in supervised OSIL we denote a set of of tasks as $\mathbb{T}$, each individual task $\bm{t} \in \mathbb{T}$, and a distribution of demonstrations of task $\bm{t}$ as $\mathbb{D}(\bm{t})$. 
The supervised OSIL objective is to train a policy which, conditioned on a demonstration $d \sim \mathbb{D}(\bm{t})$, can accomplish a task $\bm{t}$.
This amounts to learning a goal conditioned policy $\pi_\theta(a_t | s_t, d)$, parameterized by $\theta$, that takes an expert demonstration and the current state of the environment as input and emits the proper actions at each time-step $t$ (we differentiate time $t$ and task $\bm{t}$, which is in bold).
During training, we have access to a large dataset of demonstrations $d^{train} \sim \mathbb{D}(\bm{t}_i^{train})$, for a set of training tasks $\bm{t}_i^{train} \in \mathbb{T}^{train} \subset \mathbb{T}$, where $\bm{t}_i$ is the $i^{th}$ task.
We formulate the dataset $\mathcal{D}$ as follows

\begin{equation}
    \mathcal{D} = \{(\bm{t}_i, \{d_1^i, d_2^i, ...\} ) \; \forall \bm{t}_i \in \mathbb{T}^{train} \},
\end{equation}

We further assume the existence of a binary valued function $R_{\bm{t}}(d)$ which indicates whether a given demonstration or policy rollout $d$ successfully accomplishes the task $\bm{t}$, which we use for evaluating our method.
At test time, the policy is provided with one new test demonstration $d^{test} \sim \mathbb{D}(\bm{t})$ that can be either be a new demonstration of a seen task (i.e. $\bm{t} \in \mathbb{T}^{train}$) or a new demonstration of an unseen task (i.e. $\bm{t} \in \mathbb{T} \setminus \mathbb{T}^{train}$).

\textbf{Semi-supervised OSIL} builds on the supervised OSIL setting, which we formulate as follows.
We similarly assume access to a small labeled dataset of demonstrations $\mathcal{D}^{labeled}$ where each demonstration has its associated task label.
We additionally assume access to a large dataset of demonstrations $\mathcal{D}^{unlabeled}$ which does not have the associated task label $\bm{t}_i$.
These datasets are defined below:
\begin{align}
    \mathcal{D}^{labeled} &= \{(\bm{t}_i, \{d_1^i, d_2^i, ...\} ) \; \forall \bm{t}_i \} \\
    \mathcal{D}^{unlabeled} &= \{d_1, d_2, ... \}
\end{align}
An effective semi-supervised method should be able to leverage both annotated and un-annotated datasets effectively to maximize the performance of the OSIL agent at test time.
\section{Method}
\label{method}
At its core, OSIL can be simply construed as two modules that are jointly optimized together: (1) an encoder network $f_\phi(d)$ which embeds demonstrated trajectories into a latent space $z$, and (2) a policy decoder $\pi_\theta(a_t|s_t, z)$ that is conditioned on the demonstration embedding and current state of the environment to output actions.
The prior state of the art work on OSIL \citep{rocky_osil, traj_osil, mandi_osil} learn both the demonstration encoder module and the policy decoder jointly by minimizing the predicted action errors on the imitated trajectory, possibly with other auxiliary losses.
This method works well when paired trajectories are abundant. In the more realistic semi-supervised OSIL setting, the question becomes \textit{``How can we group sufficently abundant demonstration pairs from the unlabeled data to train an OSIL agent?''} To address this, we propose an iterative student teacher method. 

\subsection{Student-Teacher Training}
The core of our hypothesis is that discriminating or clustering trajectories that share the same semantic task is easier (and thus more data efficient) compared to generative modeling of actions to accomplish a task. To instantiate this in practice, we use a teacher-student self-training paradigm \citep{xie2020self} to effectively remove the need for large human-annotation on task labels.
In our setting, a "teacher" is the encoder $f_\phi$ that embeds trajectories into the latent space. Using a quality teacher encoder, we can retrieve the $k$-nearest neighbors of each trajectory in the dataset using a distance measure (e.g L2 distance) on the embedding space and use that as a labeled pair for downstream training of a student OSIL policy. 

To train the teacher encoder, we proceed with the standard OSIL training procedure on the smaller labeled dataset, $\mathcal{D}^{labeled}$.
The encoder and policy are trained end to end with an imitation loss on the predicted action from the policy, $\pi_{\theta}(a_t | s_t, z)$, where $z = f_\phi(d_{\bm{t}})$.
To encourage learning a more structured latent space, we also employ a contrastive InfoNCE loss \citep{info_nce}, where a positive pair is taken from the labeled subset of data, and the rest of the goals in the batch are treated as negative examples.
This structured latent space is necessary for teacher relabelling. 
In general, we also find that the contrastive loss helps with learning a better OSIL policy with higher task success rate, which is consistent with the works of \citet{James2018TaskEmbeddedCN, mandi_osil}.

\begin{figure}
  \begin{subfigure}{0.49\textwidth}
    \includegraphics[width=\linewidth]{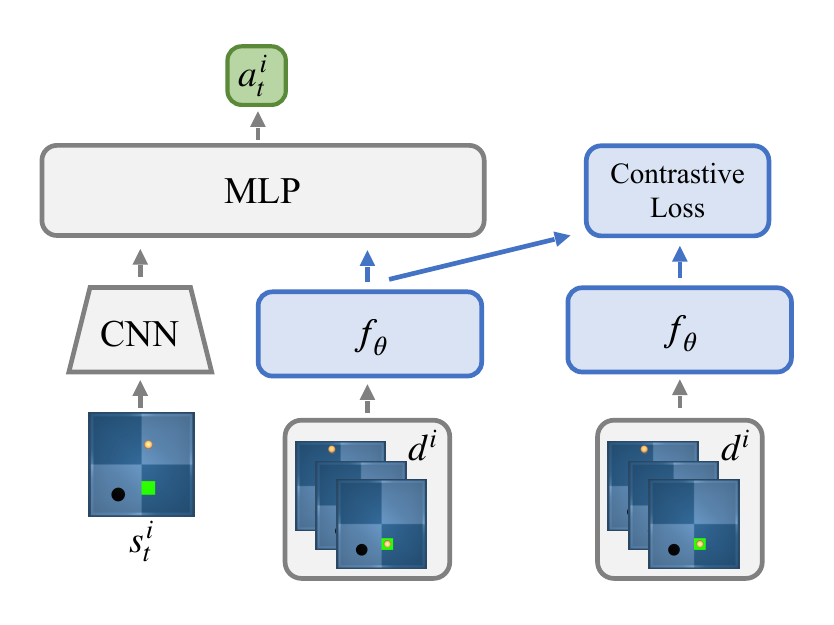}
    \caption{Architecture overview} \label{fig:architexture_overview}
  \end{subfigure}%
  \hspace*{\fill}   
  \begin{subfigure}{0.42\textwidth}
    \includegraphics[width=\linewidth]{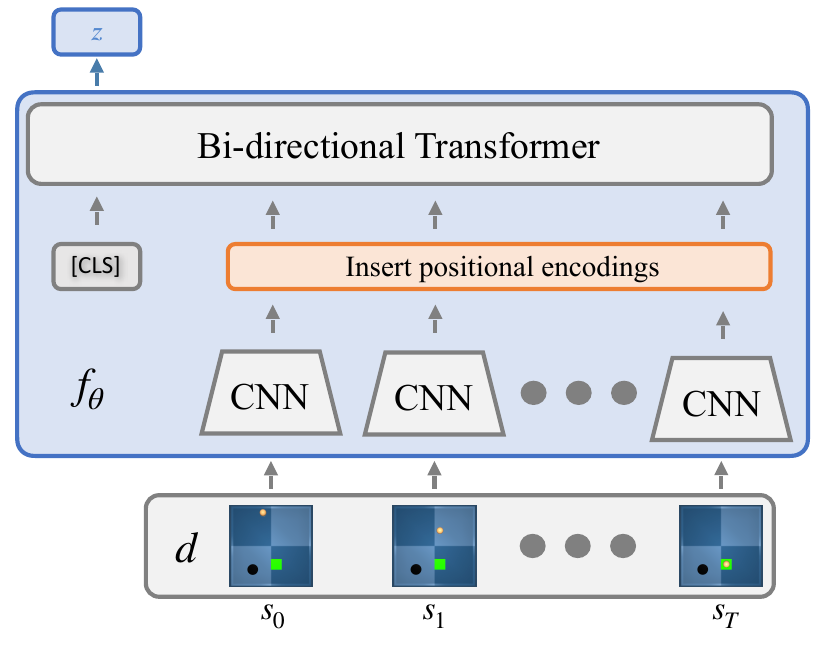}
    \caption{Trajectory Encoder} \label{fig:trajectory_encoder}
  \end{subfigure}%
  \hspace*{\fill}   

\caption{The architecture used in our algorithm. (a) shows the generic structure of a OSIL agent, which consists of a generic demonstration encoder $f_\phi$ and the $\pi_\theta(a_t|s_t, z)$ task latent conditioned policy, which comprises of image encoder.. (b) shows one potential instantiation of the demonstration encoder, which leverages a bi direction transformer to encode the trajectory. This is used for the pinpad sequential navigation task, which requires reasoning over the entire trajectory.} \label{fig:1}
\vspace*{-20pt}
\end{figure}

After training the teacher encoder to convergence, we then generate a set of pseudo labels for the trajectories in the unlabeled dataset.
This is done by embedding all of the demonstrations of the dataset $\mathcal{D}^{unlabeled}$ with the teacher encoder $f_{\phi}$. 
We then find the $k$ nearest neighbors of each demonstration in the embedding space, where $k$ is a hyperparameter.
Let $k\text{NN}_{\phi}(d, \mathcal{D})$ denote the $k$ nearest neighbors of $d$ in the dataset $\mathcal{D}$ using the feature embeddings from a demonstration encoder $f_{\phi}$.
If the nearest neighbors are demonstrations associated with the same semantic task, we can supervise an effective student OSIL policy with this dataset of pseudo-pairs of trajectories, which we formulate as:
\begin{equation}
    \mathcal{D}^{pseudo\_labeled} = \{(d_i, \{k\text{NN}_{\phi}(d_i, \mathcal{D}^{unlabeled})\}) \; \forall d_i \in \mathcal{D}^{unlabeled} \}
\end{equation}

Finally, the student policy is trained using both $\mathcal{D}^{pseudo\_labeled}$ and $\mathcal{D}^{labeled}$.
During training we continue to use the labeled dataset for the imitation and contrastive losses, but additionally sample batches from the pseudo-labeled dataset, which is trained only with the imitation loss.
We can continue iterating this process by treating the encoder $f_\phi$ of the trained student as the teacher for the subsequent round and improving the accuracy on the KNN retrievals from the unlabeled dataset until we get diminishing returns from the process.  

\subsection{Architecture}

An overview of the architecture is shown in Figure \ref{fig:architexture_overview}. 
We use the same architecture for teacher and student with same number of parameters.
In general, the demonstration encoder $f_{\phi}$ is flexible and can take any form, but should be expressive enough to learn meaningful representations of the demonstration trajectories.
Following conditional policies \citep{jang2022bc}, we utilize an MLP policy which takes the demonstration embedding z through FiLM conditioning \citep{perez2018film}. 
The focus of our work is on the procedure of making OSIL more data efficient. We therefore do not consider more complex encoder decoder architectures, for which we refer to prior work.
In this work we also focus our experiments on visual imitation, for which we use a CNN encoder to obtain frame-level visual representations of 64x64 images with a simple 5 layer CNN.

For many OSIL tasks, the final frame is enough to specify the desired intent, which we find true for this environment.
A commonly used strategy is to form a summary of the demonstration trajectory by taking a few key frames \citep{rocky_osil, James2018TaskEmbeddedCN}.
For these tasks (e.g. goal reaching) we simply use the final frame image embedding as the representation of the task.
Figure \ref{fig:trajectory_encoder} on the other hand, illustrates a more general solution to embed the entire demonstrated trajectory. In this model, we treat the embedding of each frame as a separate token and use a bi-directional transformer to learn the task encoding. The transformer model has the capacity to learn which frames are important to fully describe the task. Refer to Appendix \ref{append_hyper} for more hyperparameter details.
\section{Experiments}
\label{experiments}
Through our experiments, we aim to study the effectiveness of the semi-supervised OSIL setting, as well as the performance of our algorithm. Concretely, we study the following questions.
\vspace{-5pt}
\begin{itemize}
    \itemsep0em
    \item How to train the demonstration encoder to effectively cluster trajectories?
    \item How to use the learned clusters to effectively improve agent performance?
\end{itemize}
\vspace{-10pt}

\subsection{Environment setup}

\begin{figure}[t!]
\centering
\begin{subfigure}{0.43\textwidth}
    \includegraphics[width=\textwidth]{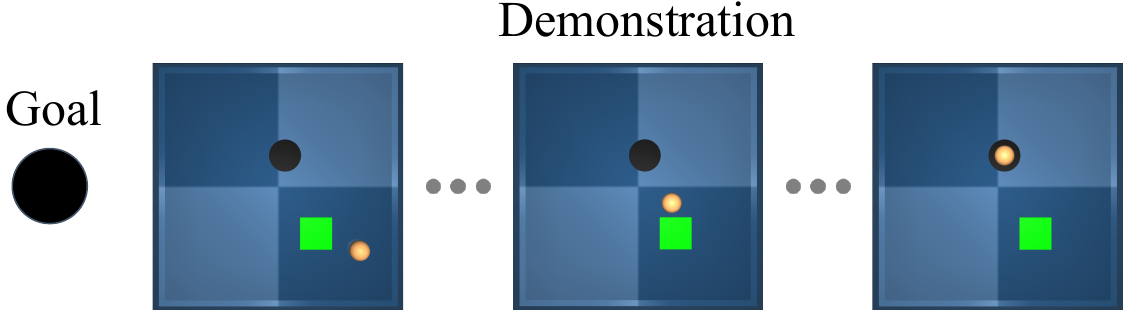}
    \caption{Semantic Goal Navigation Environment.}
    \label{fig:first}
\end{subfigure}
\hfill
\begin{subfigure}{0.43\textwidth}
    \includegraphics[width=\textwidth]{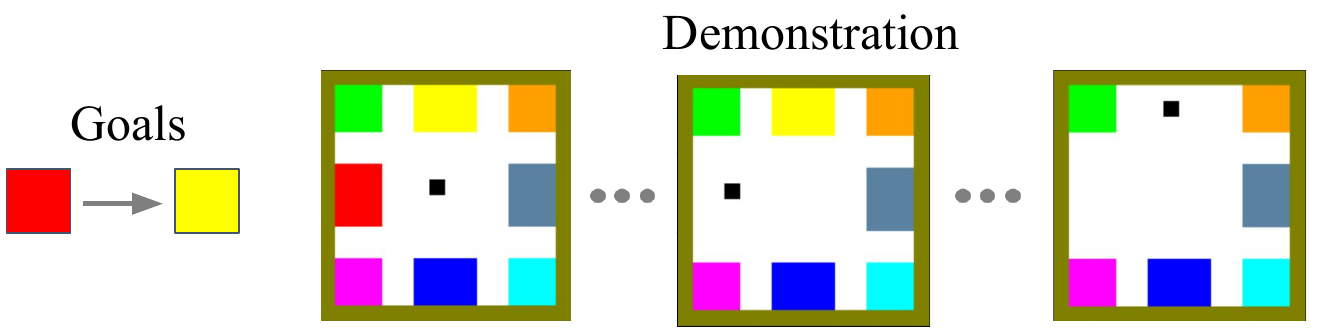}
    \caption{Sequential Goal Navigation Environment.}
    \label{fig:second}
\end{subfigure}
\hfill
\caption{Sample goals and corresponding demonstration visualizations for the two tasks.}
\label{fig:env_illustration}
\end{figure}

\paragraph{Semantic Goal Navigation.} We construct a custom  pointmass-based reaching task using the MuJoCo simulator \citep{mujoco} with the DMControl suite \citep{dm_control}. This task is inspired from the simulated reaching task first introduced in \citep{finn_meta_osil}.
The task is to navigate the pointmass to a goal of a given color and shape when also presented with a distractor goal of a different color and shape.
Concretely, there are 2 shapes and 5 possible colors the shapes can take on, totalling 10 variations for each object, and 100 possible semantic scenes.
See Figure~\ref{fig:env_illustration} for a visual illustration. Note that within each scene configuration, the locations of the objects can be randomized.
We collect 800 trajectories for each target goal object, resulting in a total training dataset size of 8000 trajectories. 

\paragraph{Sequential Goal Navigation.} We use a modified version of the discrete pinpad world environment from \citet{hafner2022deep}. This task requires the agent to navigate and press two buttons out of six in a specified order. The agent is only considered successful if it is able to correctly reach all the goals in the correct sequence.
There are 6 possible goal pads for the agent to reach, totaling 30 tasks. The agent’s action is one of five possible actions: up, down, left, right, or no-op. The observation space is the raw pixels in the environment.
See Figure~\ref{fig:env_illustration} for a visual illustration.
We randomize both the color assignments of the pads and the agent starting location for each task variation.
The agent must pay attention to the entire trajectory to correctly determine the desired task. As such, we parametrize the demonstration encoder for this environment as a small bi-directional transformer that takes in a sequence of states and a class token to predict a latent $z$ encoding of the trajectory. 

\paragraph{Dataset Collection.}
We employ a scripted policy to collect demonstrations for each task variation. Specifically, we reset the initial state of the environment and agent randomly, then run the scripted expert policy. During training we limit the number of demonstrations per each task that the agent gets to see for supervision in order to create a semi-supervised scenario. However, we use the entire collected dataset as a large pool of unlabeled expert trajectories during training. We evaluate our method on two environments described above.

\subsection{Metrics}

\paragraph{Task Success.} Our goal is to maximize task success rate using limited task labeled demonstrations. 
For both environments, we report the success rate of the agent as the performance after 100 trials in the environment, averaged over 3 seeds. We evaluate the agent on both new instantiations of the training tasks and an unseen test task, which we report as "Train" and "Test" respectively. We use different numbers of the total labeled trajectories to show how the number of labeled trajectories effects final task performance.

\paragraph{Trajectory Retrieval (TR) Score} For each trajectory in $d^{\text{test}} \in \mathcal{D}$, we retrieve the $K$ nearest neighbors by measuring the L2 distance in the embedding space of the teacher.
Let $d_{i}^{\text{ret}}$ be the $i^{th}$ retrieved trajectory and $\bm{t}$ be  the task label of $d^{test}$. 
For each trajectory, the retrieval accuracy is defined as the percentage of time that $R_{\bm{t}}( d^{\text{ret}}_i) = 1$. 
We take an average of this measure across all samples in the training set. 

\begin{equation}
    TR_{score}(\mathcal{D}) = \frac{1}{|\mathcal{D}|} \sum_{d\in\mathcal{D}} 
    \frac{1}{k}\sum_{j=1}^k R_{\bm{t}}( d^{\text{ret}}_j)
\end{equation}

\subsection{Results}
For each experiment, we train the OSIL policy using the learned goal embedding and behavior cloning loss on the labeled subset of data. We report the task success rate and trajectory retrieval scores for all experiments. 

\paragraph{Semantic Goal Navigation}
First we consider the Semantic Goal Navigation pointmass task. We consider 5 main settings:
\vspace{-5pt}
\begin{enumerate}
    \itemsep0em
    \item An agent trained with only the imitation loss on the demonstrated actions.
    \item An agent trained with an additional contrastive loss on the goal embeddings in addition to to the imitation loss.
    \item The same as (2.) but with an added self-supervised loss on the entire dataset (including unlabeled data).
    \item A student model trained by using the demonstration encoder (2.) as a teacher model.
    \item A student model trained by using the demonstration encoder of (3.) as a teacher model.
\end{enumerate}

The model trained with the method specified in (3.) acts as an alternative semi-supervised baseline in the special case of using the final frame as the demonstration representation. In this setting, we use the supervised labels as in the supervised OSIL case, but further leverage the unlabeled trajectories through adding an additional self supervised loss contrastive loss on augmentations of the goal image \citep{info_nce}. The augmentations we use are restricted to random flip and random crop.

\begin{figure}[t!]
\centering
\begin{subfigure}{\textwidth}
    \includegraphics[width=\textwidth]{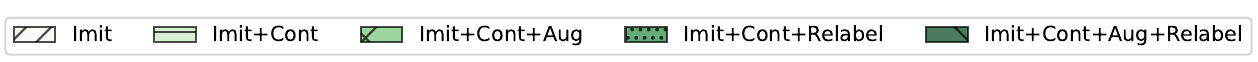}
\end{subfigure}
\hfill
\begin{subfigure}{0.49\textwidth}
    \includegraphics[width=\textwidth]{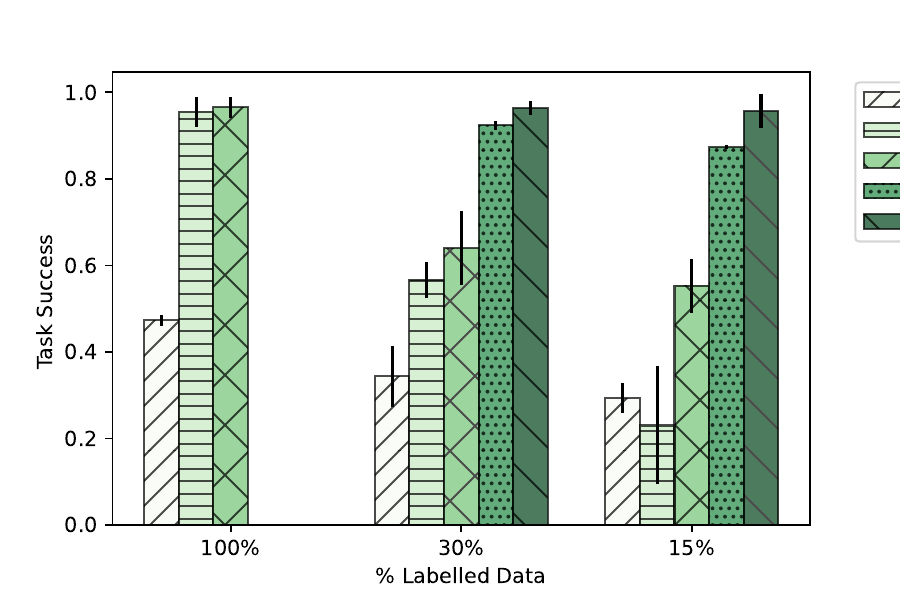}
    \caption{Train success rates.}
    \label{fig:first}
\end{subfigure}
\hfill
\begin{subfigure}{0.49\textwidth}
    \includegraphics[width=\textwidth]{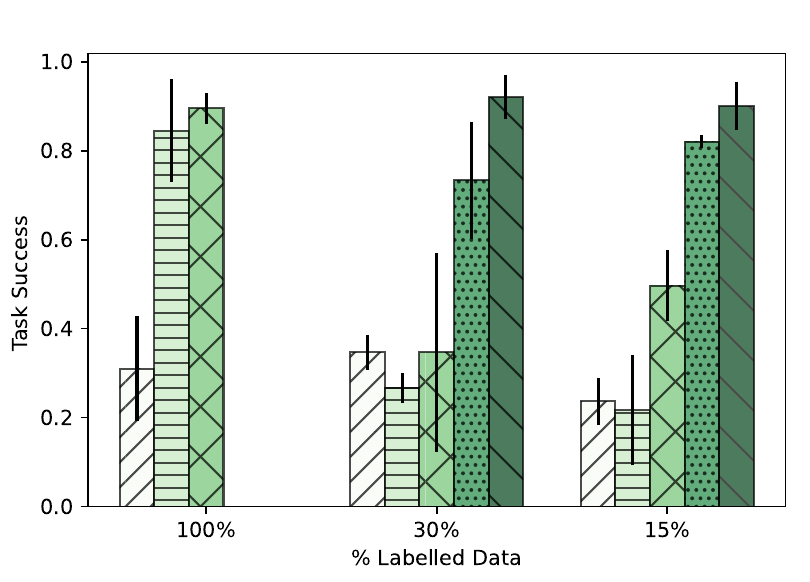}
    \caption{Test success rates.}
    \label{fig:second}
\end{subfigure}
\hfill
\caption{Task success rates for the Semantic Goal Navigation Task.}
\label{fig:ffreacherbarplot}
\end{figure}

\begin{table}
  \small
  \caption{Trajectory Retrieval: Final Frame Semantic Goal Navigation}
  \label{pointmass_results}
  \centering
  \begin{tabular}{lllllll}
    \toprule
    &  & \multicolumn{5}{c}{Retrieval \% with $k$=}\\
 
    \cmidrule(r){3-7}
    \% Labeled Data        & Method                   & 1    & 10    & 50    & 100   & 200      \\
    \midrule                                                         
    \multirow{3}{*}{100\%} & Imitation                & 11.3    & 11.    & 10.7  & 10.6  & 10.5 \\
                           & +Contrastive             & 90.9    & 91     & 91.2  & 90.9  & 90.7 \\
                           & +Contrastive+Aug         & 93.5    & 92.5   & 91.7  & 91.3  & 90.8 \\
    \midrule        
    \multirow{3}{*}{30\%}  & Imitation                & 11.8 & 11.8    & 11.5 & 11.4  & 11.3  \\
                           & +Contrastive             & 88.8 & 88.5   & 88.1 & 87.90 & 87.8  \\
                           & +Contrastive+Aug         & 93.7 & 92.9  & 91.9  & 91.4  & 90.8 \\
                           & +Contrastive+Relabel     & 91.1 & 90.6  & 90.3  & 89.9  & 89.6 \\
                           & +Contrastive+Aug+Relabel & 93.7 & 92.5  & 91.2  & 90.7  & 90.3 \\
    \midrule        
    \multirow{3}{*}{15\%}  & Imitation                & 11.6 & 11.2  & 11.0  & 10.9 & 10.8 \\
                           & +Contrastive             & 63.6 & 62.8  & 61.6  & 61.  & 60.   \\
                           & +Contrastive+Aug         & 91.6 & 90.6  & 90.   & 89.9 & 89.7  \\
                           & +Contrastive+Relabel     & 74.3 & 73.2  & 71.9  & 71.2 & 70.4  \\
                           & +Contrastive+Aug+Relabel & 91.8 & 90.7  & 90.1  & 89.9 & 89.6  \\
    \bottomrule
  \end{tabular}
  \label{tab:ffreacher_retrieval}
\end{table}

\begin{figure}
    \centering
    \includegraphics[width=0.99\linewidth]{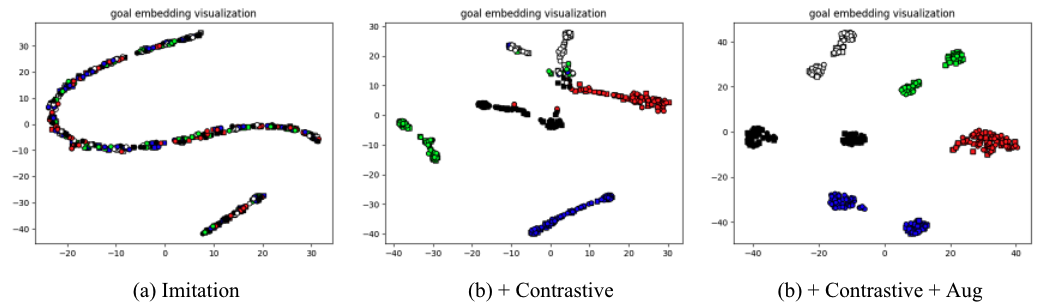}
    \caption{TSNE visualizations of the learned embeddings where the only 15\% of the dataset is labeled. (a) shows the embedding trained with imitation loss only. (b) adds the contrastive loss on the labeled subset of data (c) additionally adds a self supervised loss with images augmentations.}
    \label{cluster_vis}
\end{figure}

Figure \ref{fig:ffreacherbarplot} show the task performance across each experiment.
As expected, train and validation performance drops when the amount of labeled data decreases. 
Without relabelling, we see some task performance gains from applying both the contrastive loss and self unsupervised losses. 
After training a student model using the pseudo-labels from the representations learned by the teacher encoder, we see a leap in performance, matching an agent that has access to 100\% ground truth labels, even when only using 15\% of the labels. 

Table \ref{tab:ffreacher_retrieval} shows the trajectory retrieval scores across different values of $k$.
Despite having much less labeled data and decreased task performance, the retrieval scores consistently remain high.
This suggests that we are able to learn a meaningful representation for the task, allowing us to cluster the trajectories. We find that the contrastive loss is necessary for learning representations that have high retrieval. Interestingly, we find that relabeling gives much greater gains for task success while augmentation gives more benefit for retrieval, which supports the hypothesis that relabeling gives datapoints for the OSIL training and the contrastive loss (which typically relies on augmented views of data points) helps representation learning, but in a way thats not directly optimizing for the task objective. We show 2D visualizations of the learned embedding in Figure \ref{cluster_vis}. We additionally explore the effect of the the number of possible pairs $k$ in Appendix \ref{append_k}.

\paragraph{Sequential Goal Navigation}
Next we examine a task which requires the trajectory encoder to learn a time-dependent encoding of the trajectory, rather than having the task fully specified by the final frame.
For this we employ the more general trajectory demonstration encoder shown in Figure \ref{fig:trajectory_encoder}.
Similarly, we see in Table \ref{pinpad_results} that our teacher-student relabeling method allows the agent to improve task performance and almost match the agent trained with a fully labeled dataset, even in much lower labeled data regimes (5\%). This suggests that the teacher encoder is able to pay attention to the temporal nature of the demonstrations and generate effective pseudo-labels. Similar to the semantic navigation task, the learned encoder maintains a high trajectory retrieval score across  different choices of $k$.

\begin{table}
  \small
  \caption{Task Success Results: PinPad}
  \label{pinpad_results}
  \centering
  \begin{tabular}{llll}
    \toprule
    &  &\multicolumn{2}{c}{Task Success \%}  \\  \cmidrule(r){3-4}
        \% Labeled   & Method                   & Train  & Val \\
    \midrule         
    100\%                  & +Contrastive             & $92.3\pm2.9$      & $41.7\pm15.1$    \\
    \midrule        
    \multirow{2}{*}{10\%}  & +Contrastive             & $70.7\pm4.5$       & $18.3\pm3.3$    \\
                          & +Contrastive+Relabel     &   $84\pm4.6$     & $36.3\pm10.6$     \\
    \midrule        
    \multirow{2}{*}{5\%}  & +Contrastive             & $49\pm1.4$       &  $7.7\pm1.6$    \\
                          & +Contrastive+Relabel     & $82.3\pm10$       & $34.3\pm22.6$    \\
    \bottomrule
  \end{tabular}
\end{table}

\section{Conclusions}
\label{sec:conclusions}
In this paper, we introduce the problem setting of semi-supervised OSIL, which we believe to be a more realistic setting for developing OSIL methods that can scale to real world settings. In semi-supervised OSIL we aim to maximize agent performance in settings where we have access to a large set of task-agnostic expert demonstrations,  but only a small task-labeled dataset.
We introduce a student teacher training method and show that training a teacher network based on the limited labeled data and bootstrapping on the resulting task encoder can allow us to assign effective pseudo-labels to the large unlabeled dataset.
Using the pseudo-labeled dataset to train a student network can result in out-performing its teacher, reaching task performance parity with a model trained on much more labeled data. We evaluate our methodology on simulated environments with varying complexity and showed that this can be a promising direction towards semi-supervised OSIL. 

Our work aims to provide agents the ability to quickly imitate a demonstration. The work does not assume any particular type of demonstration. A malicious actor might be able to provide nefarious demonstrations to AI agents and safeguards must be considered when deploying such imitation learning systems in the real world. At a more immediate level, we do not anticipate any societal risks due to this work.

\bibliography{references}
\bibliographystyle{rlc}

\newpage
\appendix
\section{Appendix}

\subsection{Hyperparameters}
\label{append_hyper}

In both experiments the image observations are first encoded with a 5 layer CNN with ReLU activiations. The CNN encoder is shared across embedding the current state and the demonstration trajectory. The policy network is also the same, which consists of 3 FiLM blocks using GELU nonlinearities and using 128 hidden units per layer. For all experiments that use the contrastive objective across paired trajectories, a weighting of 10 is used on the contrastive loss.
For all experiments using pseudo-labeled trajectories, a weighting of 0.5 is used on the imitation loss with pseudo-labeled trajectories.

\paragraph{Semantic Goal Navigation}
For this task we adopt the oracle trajectory encoder, which takes the final frame of the trajectory (which fully specifies the task) and encodes it with CNN. The images are 64x64.
The policy is trained with a learning rate of 1e-3 with 4000 warm up steps.Frame stacking of 2 is employed on the observations. For each experiment we train for 200k iterations.

For the self supervised augmentations, we employ random resizing, cropping, horizontal flip, and vertical flip.
An additional one layer projection is applied before applying the self supervised contrastive loss, which we employ with a weight of 0.05.

\paragraph{Sequential Goal Navigation}
For this task, we make no assumptions on what frames are important and use a bidirectional transformer that attends over all states in the trajectory.
The transformer has 2 hidden layers and 2 attention heads, and the goal encoding is extracted with an additional class token.
Images are 16x16. We use a learning rate of 3e-4 and train for 60k iterations.

\subsection{Effect of more pseudo label pair value $k$}
\label{append_k}

Here we study how choosing different values of $k$ and different iterations of relabeling effect final performance.
In the pseudo-labeling stage, we fix $k$ controls the number of possible pairs each unlabeled trajectory can use for training. In addition, we experiment with using the student model as a teacher model for one additional iteration of training. We use the Semantic Goal reaching task, with 15\% of the full dataset size. The results are summarized in Table \ref{multi_iteration}.
Results are mixed overall, and it seems the exact choice of $k$ does not have a significant impact on results. In addition it seems that repeating the pseudo labeling process for additional iterations does not have any significant gains on performance. This could be due to the simplicity of the task, as well as the already high trajectory retrieval scores across all $k$. We suspect that for more difficult tasks, these parameters will have a more significant impact on final performance.

\begin{table} [h]
  \caption{Iterative Relabeling on Semantic Goal Navigation}
  \label{multi_iteration}
  \centering
  \begin{tabular}{lllll}
    \toprule
    & \multicolumn{4}{c}{Test Success rate \% with $k$=}\\
    \cmidrule(r){2-5}
    Iteration & 10 & 50 & 100 & 200 \\
    \midrule                                                                 
    1         & $82 \pm 1.4$  & $88.7\pm6.9$  & $83.3\pm17.4$   & $95.4\pm1.2$   \\
    2         & $88\pm2.1$  & $87.5\pm4.5$  & $78.5\pm2.5$   & $86.5\pm0.5$      \\
    \bottomrule
  \end{tabular}
\end{table}

\end{document}